\documentclass{article}
\usepackage{spconf}

\usepackage{xcolor} 

\usepackage{cite} 

\usepackage[pdftex]{graphicx}
\graphicspath{{../figures/}}
\DeclareGraphicsExtensions{.pdf} 

\usepackage{amsmath}
\usepackage{amsthm}
\usepackage{amsfonts}
\usepackage{amssymb}

\usepackage{array} 

\usepackage[caption=false,font=footnotesize]{subfig}

\usepackage{stfloats} 

\usepackage{url} 

\makeatletter
    \@ifpackageloaded{xcolor}{}{\usepackage{xcolor}}
\makeatother
\makeatletter
    \@ifpackageloaded{needspace}{}{\usepackage{needspace}}
\makeatother

\newcommand{\uppercaseGreek}[1]{%
    \begingroup
    \ucmathlist\MakeUppercase{#1}%
    \endgroup
}

\newcommand{\ucmathlist}{%
    \def\alpha{\mathrm{A}}%
    \def\beta{\mathrm{B}}%
    \let\gamma=\Gamma
    \let\delta=\Delta
    \def\epsilon{\mathrm{E}}%
    \def\varepsilon{\mathrm{E}}%
    \def\zeta{\mathrm{Z}}%
    \def\eta{\mathrm{H}}%
    \let\theta=\Theta
    \let\vartheta=\Theta
    \def\iota{\mathrm{I}}%
    \def\kappa{\mathrm{K}}%
    \let\lambda=\Lambda
    \def\mu{\mathrm{M}}%
    \def\nu{\mathrm{N}}%
    \let\xi=\Xi
    \let\pi=\Pi
    \let\varpi=\Pi
    \def\rho{\mathrm{P}}%
    \def\varrho{\mathrm{P}}%
    \let\sigma=\Sigma
    \def\tau{\mathrm{T}}%
    \let\upsilon=\Upsilon
    \let\phi=\Phi
    \let\varphi=\Phi
    \def\chi{\mathrm{X}}%
    \let\psi=\Psi
    \let\omega=\Omega
}

\makeatletter
    \@ifpackageloaded{amsthm}{}{

        \usepackage{amsthm}
    }
\makeatother
\theoremstyle{plain}

\theoremstyle{definition}

\makeatletter
\def\renewtheorem#1{%
    \expandafter\let\csname#1\endcsname\relax
    \expandafter\let\csname c@#1\endcsname\relax
    \gdef\renewtheorem@envname{#1}
    \renewtheorem@secpar
}
\def\renewtheorem@secpar{\@ifnextchar[{\renewtheorem@numberedlike}{\renewtheorem@nonumberedlike}}
\def\renewtheorem@numberedlike[#1]#2{\newtheorem{\renewtheorem@envname}[#1]{#2}}
\def\renewtheorem@nonumberedlike#1{  
    \def\renewtheorem@caption{#1}
    \edef\renewtheorem@nowithin{\noexpand\newtheorem{\renewtheorem@envname}{\renewtheorem@caption}}
    \renewtheorem@thirdpar
}
\def\renewtheorem@thirdpar{\@ifnextchar[{\renewtheorem@within}{\renewtheorem@nowithin}}
\def\renewtheorem@within[#1]{\renewtheorem@nowithin[#1]}
\makeatother
\input{auxFiles/symbolDef.sty}
\input{auxFiles/riceColors.sty}

\usepackage{textcomp} 
\usepackage{IEEEtrantools} 
\usepackage{balance} 

\def\intinfty{\int_{-\infty}^{+\infty}}

\title{UNROLLING PARTICLES: UNSUPERVISED LEARNING OF SAMPLING DISTRIBUTIONS}

\name{Fernando Gama, Nicolas Zilberstein, Richard G. Baraniuk, Santiago Segarra\thanks{This work was partially supported by NSF under award CCF-2008555. Email: \{fgama,nzilberstein,richb,segarra\}@rice.edu}}
\address{Department of Electrical and Computer Engineering,
         Rice University,
         Houston, TX}

\begin{document}


\ninept


\maketitle


\begin{abstract}
Particle filtering is used to compute good nonlinear estimates of complex systems. It samples trajectories from a chosen distribution and computes the estimate as a weighted average. Easy-to-sample distributions often lead to degenerate samples where only one trajectory carries all the weight, negatively affecting the resulting performance of the estimate. While much research has been done on the design of appropriate sampling distributions that would lead to controlled degeneracy, in this paper our objective is to \emph{learn} sampling distributions. Leveraging the framework of algorithm unrolling, we model the sampling distribution as a multivariate normal, and we use neural networks to learn both the mean and the covariance. We carry out unsupervised training of the model to minimize weight degeneracy, relying only on the observed measurements of the system. We show in simulations that the resulting particle filter yields good estimates in a wide range of scenarios.
\end{abstract}


\begin{keywords}
algorithm unrolling, particle filtering, unsupervised learning
\end{keywords}


\section{Introduction} \label{sec:intro}



Particle filtering is a signal processing technique typically used to compute nonlinear estimates of system states based on available measurements. It consists in sampling system trajectories, also known as particles, and leveraging some form of the law of large numbers to produce the estimate \cite{Doucet2000-ParticleFiltering, Djuric2003-ParticleFiltering, Godsill2019-ParticleFiltering}. Particle filtering has found widespread use in tracking \cite{Gong2021-ParticleTracking}, robotics \cite{Chisci2021-ParticleRobotics}, and communications \cite{Shrivastava2021-ParticleCommunication}, and novel uses in medicine \cite{Shakir2021-ParticleMedicine} and contact tracing \cite{Tu2021-ParticleContactTracing}.

Particle filtering is able to obtain good estimates of complicated nonlinear systems in a computationally tractable way. Ideally, one would like to sample particles directly from the posterior distribution of the states given the measurements, so that one can obtain an unbiased estimator that is known to converge as more particles are considered. Unfortunately, sampling from this posterior distribution is oftentimes intractable or outright impossible.

Particle filtering addresses this issue by sampling trajectories, instead, from some other distribution from which it is easy to sample. By doing so, the resulting estimate becomes biased, although asymptotically unbiased and convergent under some mild conditions \cite{Doucet2000-ParticleFiltering}. More importantly, the estimate now becomes a weighed average of the particles, where the weights are dependent on the chosen sampling distribution.
One major drawback of choosing a sampling distribution, is that the resulting weights tend to degenerate with the length of the trajectory, leading to a single particle carrying all the weight while the rest become zero. While this is computationally inefficient (as particles that carry no weight are still computed), it predominantly affects the reliability and performance of the estimate. Since only one particle is contributing to the estimate, the variance becomes too large and convergence is unattainable.

Research in particle filtering focuses on designing appropriate distributions from which it is easy to sample from, while still leading to as little degeneracy as possible. A case of particular interest is when one wants to improve on the computational efficiency by designing a distribution that can be sampled sequentially, leading to the Sequential Importance Sampling (SIS) particle filter \cite{Doucet2000-ParticleFiltering}. It can then be proved that the sampling distribution that minimizes the degeneracy is given by the posterior of the current state, given the past state and the current measurement \cite[Prop. 2]{Doucet2000-ParticleFiltering}. Unfortunately, this distribution is typically intractable. This can be addressed by using of Gaussian mixture models \cite{Rehman2018-GMM} or stochastic convex programming \cite{Ryu2014-StochasticProgramming}, among other approaches \cite{Elvira2021-AdvancesParticleFiltering}.

The objective of this work is to \emph{learn} the sampling distribution, instead of designing it. To do so, we leverage the framework of algorithm unrolling or unfolding~\cite{Balatsoukas2019-DeepUnfolding, Monga2021-AlgorithmUnrolling, Chowdhury21-UWMMSE}. By parametrizing the sampling distribution as a multivariate normal, and identifying the SIS particle filter as an iterative algorithm, we propose to learn a separate neural network for the mean at each time instant. For learning the covariance, we utilize another neural network that remains constant over time. The resulting SIS particle filter is trained in an unsupervised manner by maximizing the log-sum of the weights, looking to avoid the degeneracy of the algorithm.

There has been work on using learning to improve on particle filtering. Typically, learning is used mostly to estimate the model or transforming the variables into spaces more amenable for sampling \cite{Brock2018-ParticleDifferentiable}. Recurrent neural networks (RNNs) have also been used to learn the mean of a multivariate normal sampling distribution as well as the particle weights \cite{Karkus2020-ParticleRNN}. Most importantly, in all these cases, the neural networks are trained using supervised learning. This requires access to true trajectories of the system, which are typically unavailable, and it does not guarantee that the trajectories observed at test time will be similar enough to ensure generalization. We address this fundamental drawback by proposing a trainable particle filtering based on unsupervised learning.

\vspace{1mm}
\noindent
{\bf Contributions.} The contributions of this paper are twofold:\\
1) We propose (to the best of our knowledge) the first unsupervised learning approach to train the sampling distribution of a particle filter.
We attain this by promoting particle weights to be as uniform as possible, thus avoiding degeneracy.\\
2) We illustrate the superior performance of the learnable particle filter compared with classical benchmarks through numerical experiments in linear and nonlinear settings.


\section{Particle Filtering} \label{sec:particle}



Let $\{\vcx_{t}\}_{t \geq 0}$ be a sequence of states $\vcx_{t} \in \fdR^{N}$ that describe a given system. These states are considered unobservable. Instead, we have access to a sequence of measurements $\{\vcy_{t}\}_{t \geq 0}$ with $\vcy_{t} \in \fdR^{M}$. The states and measurements are related by the following, given model
\begin{subequations} \label{eq:modelDescription}
    \begin{IEEEeqnarray}{lll}
    \text{Initial state:}
    &\quad \vcx_{0} \sim \fnp(\vcx_{0})
    \label{subeq:initialState} \\
    \text{Transition:}    &\quad \fnp(\vcx_{t}|\vcx_{t-1}),
    & \quad t=1,2,\ldots
    \label{subeq:transition} \\
    \text{Measurement:}   &\quad \fnp(\vcy_{t}|\vcx_{t}),
    & \quad t=0,1,\ldots.
    \label{subeq:measurement}
    \end{IEEEeqnarray}
\end{subequations}
This is a hidden Markov model (HMM) and is typically used in many applications \cite{Doucet2000-ParticleFiltering, Djuric2003-ParticleFiltering, Godsill2019-ParticleFiltering, Gong2021-ParticleTracking, Chisci2021-ParticleRobotics, Shrivastava2021-ParticleCommunication, Shakir2021-ParticleMedicine, Tu2021-ParticleContactTracing}.

In this context, the broad objective is to estimate the sequence of states $\{\vcx_{t}\}$ by relying on the measurements $\{\vcy_{t}\}$. The most general of such estimates is the posterior probability $\fnp(\vcx_{0:t}|\vcy_{0:t})$, where $\vcx_{0:t}$ denotes either the set $\{\vcx_{\tau}\}_{\tau=0,\ldots,t}$ or the vector $[\vcx_{0}^{\Tr}, \vcx_{1}^{\Tr},\ldots,\vcx_{t}^{\Tr}]^{\Tr} \in \fdR^{N(t+1)}$ depending on context; analogously for $\vcy_{0:t}$. The posterior $\fnp(\vcx_{0:t}|\vcy_{0:t})$ can be computed recursively as
\begin{equation} \label{eq:posterior}
    \fnp(\vcx_{0:t}|\vcy_{0:t})
        = \frac{\fnp(\vcy_{t}|\vcx_{t})\fnp(\vcx_{t}|\vcx_{t-1})}{\fnp(\vcy_{t}|\vcy_{0:t-1})} \fnp(\vcx_{0:t-1}|\vcy_{0:t-1}).
\end{equation}
Typically, however, we are interested in computing summaries of $\fnp(\vcx_{0:t}|\vcy_{0:t})$ such as the mean squared error estimate $\xp[\vcx_{0:t}|\vcy_{0:t}]$, the posterior  of the current time instant $\fnp(\vcx_{t}|\vcy_{0:t})$ or a one-step prediction $\fnp(\vcx_{t+1}|\vcy_{0:t})$, among many others \cite{Doucet2000-ParticleFiltering, Djuric2003-ParticleFiltering, Godsill2019-ParticleFiltering}.

All of these estimates involve, in one form or another, an integral over the the posterior density. We denote such integral as follows
\begin{equation} \label{eq:integral}
    I(\fnf_{t}) = \xp \big[ \fnf_{t}(\vcx_{0:t}) |\vcy_{0:t} \big] = \intinfty \fnf_{t}(\vcx_{0:t}) \fnp(\vcx_{0:t}|\vcy_{0:t})\ d\vcx_{0:t},
\end{equation}
for some function $\fnf_{t}:\fdR^{N} \times \cdots \times \fdR^{N} \to \fdR^{O}$, where $O$ is a prespecified output dimension. The integral $I(\fnf_{t})$ acts as a stand-in for any of the desired summaries of the states, given the measurements, as described before. We note that computing such integral is usually intractable due to the high-dimensionality of the space, even if $\fnp(\vcx_{0:t}|\vcy_{0:t})$ is known in closed form.

One straightforward way of approximating $I(\fnf_{t})$ is to leverage the law of large numbers so that
\begin{equation} \label{eq:LLN}
    I_{K}^{\opt}(\fnf_{t}) = \frac{1}{K} \sum_{k=1}^{K} \fnf_{t}(\vcx_{0:t}^{(k)})
\end{equation}
converges almost surely to $I(\fnf_{t})$ as $K \to \infty$ \cite[Ch. 2]{Durrett2010-Probability}. For this to be true, the samples of the trajectory $\{\vcx_{0:t}^{(k)}\}_{k=1,\ldots,K}$ (commonly known as particles) have to be drawn from the posterior distribution, i.e., $\vcx_{0:t}^{(k)} \sim \fnp(\vcx_{0:t}|\vcy_{0:t})$. However, generating samples from the posterior distribution is usually intractable, and thus \eqref{eq:LLN} cannot be used as an estimate of $I(\fnf_{t})$.

A particle filter addresses this issue by sampling from some other distribution $\fnpi(\vcx_{0:t}|\vcy_{0:t})$ and computing the estimate as
\begin{equation} \label{eq:particleFilter}
    \schI_{K}(\fnf_{t}) = \sum_{k=1}^{K} \fnf_{t}(\vcx_{0:t}^{(k)}) \sctw_{t}^{(k)},
\end{equation}
where $\sctw_{t}^{(k)} = w_{t}^{\opt(k)}/\sum_{k=1}^{K} w_{t}^{\opt(k)}$ are the normalized weights given by $w_{t}^{\opt(k)} = \fnp(\vcy_{0:t}|\vcx_{0:t}^{(k)}) \fnp(\vcx_{0:t}^{(k)})/\fnpi(\vcx_{0:t}|\vcy_{0:t})$. The estimator in~\eqref{eq:particleFilter} is biased but converges to the true value of $I(\fnf_{t})$ as $K \to \infty$ under mild conditions \cite{Doucet2000-ParticleFiltering, Djuric2003-ParticleFiltering}.

Certainly, for the particle filter \eqref{eq:particleFilter} to be a practically useful estimate, it has to be straightforward to obtain particles $\vcx_{0:t}^{(k)}$ from the chosen distribution $\fnpi(\vcx_{0:t}|\vcy_{0:t})$. Also,  it has to be easy to compute the weights $w_{t}^{\opt(k)}$. In particular, we can restrict our attention to sampling distributions that can be updated iteratively as $\fnpi(\vcx_{0:t}|\vcy_{0:t}) = \fnpi(\vcx_{0:t-1}|\vcy_{0:t-1}) \fnpi(\vcx_{t}|\vcx_{0:t-1},\vcy_{0:t})$, leading to the SIS formulation \cite{Doucet2000-ParticleFiltering}. In this setting, at each time instant $t$ a new value $\vcx_{t}^{(k)}$ is sampled from $\fnpi(\vcx_{t}|\vcx_{0:t-1},\vcy_{0:t})$ and appended to the existing trajectory, so that $\vcx_{0:t}^{(k)} = \{\vcx_{0:t-1}^{(k)},\vcx_{t}^{(k)}\}$. The weights of \eqref{eq:particleFilter} are also computed iteratively as
\begin{equation} \label{eq:SISweights}
    w_{t}^{\opt(k)} = w_{t-1}^{\opt(k)} \frac{\fnp(\vcy_{t}|\vcx_{t}^{(k)})\fnp(\vcx_{t}^{(k)}|\vcx_{t-1}^{(k)})}{\fnpi(\vcx_{t}^{(k)}|\vcx_{0:t-1}^{(k)},\vcy_{t})},
\end{equation}
for $t=0,1,...$ and where $\fnp(\vcx_{0}^{(k)}|\vcx_{-1}^{(k)}) = \fnp(\vcx_{0}^{(k)})$. We note that the weights are easy to compute since the numerator is given by the model description \eqref{eq:modelDescription} while the denominator is the chosen sampling distribution.
Unfortunately, the unconditional variance of the weights in the SIS particle filter [cf. \eqref{eq:SISweights}] tends to increase over time \cite[Prop.~1]{Doucet2000-ParticleFiltering}, leading to the degeneration of the particles, i.e., out of the $K$ simulated trajectories only one of them will tend to have a weight $\sctw_{t}^{(k)}$ of $1$ while the rest will tend to $0$. This prevents the estimate in \eqref{eq:particleFilter} from averaging many particles, affecting the convergence rate and the quality of the estimator.

The sampling distribution for the SIS particle filter that minimizes the variance of the weights is given by $\fnpi(\vcx_{t}|\vcx_{0:t-1},\vcy_{t}) = \fnp(\vcx_{t}|\vcx_{t-1},\vcy_{t})$ \cite[Prop. 2]{Doucet2000-ParticleFiltering}. This distribution is guaranteed to slow down the degeneracy of the weights and thus help in the convergence of the estimate. Unfortunately, it is usually not possible to sample from it either. It does hint, however, at the idea that a good sampling distribution of a SIS particle filter depends only on the previous state and the current measurement. Thus, many designed sampling distributions assume this dependency, setting $\fnpi(\vcx_{t}|\vcx_{0:t-1},\vcy_{t}) = \fnpi(\vcx_{t}|\vcx_{t-1},\vcy_{t})$ \cite{Ryu2014-StochasticProgramming, Rehman2018-GMM, Elvira2021-AdvancesParticleFiltering}.


\section{Unrolling Particles} \label{sec:unrolling}



We propose to learn the sampling distribution $\fnpi(\vcx_{t}|\vcx_{0:t-1},\vcy_{0:t})$ in the context of an SIS particle filter [cf. \eqref{eq:particleFilter}-\eqref{eq:SISweights}], by exploiting the algorithm unrolling framework \cite{Balatsoukas2019-DeepUnfolding, Monga2021-AlgorithmUnrolling}. Akin to the optimal distribution, we adopt a sampling distribution that depends on the previous state and on the current measurement, i.e., $\fnpi(\vcx_{t}|\vcx_{0:t-1},\vcy_{0:t})=\fnpi(\vcx_{t}|\vcx_{t-1},\vcy_{t})$. In particular, we parametrize the sampling distribution with a multivariate normal
\begin{equation} \label{eq:MVNormalParam}
    \fnpi(\vcx_{t}|\vcx_{t-1},\vcy_{t}) = \stN \Big( \fnmu_{t}(\vcx_{t-1},\vcy_{t}), \fnSigma_{t}(\vcx_{t-1},\vcy_{t}) \Big).
\end{equation}
We note that, given the mean and the covariance matrix, sampling from a multivariate normal distribution is straightforward. Therefore, this is a sensible choice for a sampling distribution.

Following the framework of algorithm unrolling, instead of prespecifying $\fnmu_{t}(\cdot)$ and $\fnSigma_{t}(\cdot)$ in~\eqref{eq:MVNormalParam}, we learn these mean and covariance functions.
The learning architecture for the mean is given by a neural network $\fnmu_{t}(\vcx_{t-1},\vcy_{t}) = \mathrm{NN}_{t}^{\mu}(\vcx_{t-1},\vcy_{t})$ such that
\begin{equation} \label{eq:NN}
    \mathrm{NN}_{t}^{\mu}(\vcx_{t-1},\vcy_{t}) = \vcz_{t}^{(L_{t})} \ \text{where} \ \vcz_{t}^{(\ell)} = \fnrho_{t}(\mtA_{t}^{(\ell)} \vcz_{t}^{(\ell-1)} + \vcb_{t}^{(\ell)}),
\end{equation}
for $\ell=1,\ldots,L_{t}$. The matrix $\mtA_{t}^{(\ell)}$ is of size $N_{t}^{(\ell)} \times N_{t}^{(\ell-1)}$ and maps the $N_{t}^{(\ell-1)}$-dimensional input $\vcz_{t}^{(\ell-1)}$ into a vector of size $N_{t}^{(\ell)}$. The vector $\vcb_{t}^{(\ell)}$ is of size $N_{t}^{(\ell)}$. The activation function $\fnrho_{t}$ is typically nonlinear and acts over each entry of the affine transformation $\mtA_{t}^{(\ell)} \vcz_{t}^{(\ell-1)} + \vcb_{t}^{(\ell)}$. We note that the input to the neural network is given by $\vcz_{t}^{(0)} = [\vcx_{t-1}^{\Tr}, \vcy_{t}^{\Tr}]^{\Tr} \in \fdR^{N+M}$, the concatenation of the previous state $\vcx_{t-1}$ and the current measurement $\vcy_{t}$. This implies that $N_{t}^{(0)} = M+N$. The values of $\mtA_{t}^{(\ell)}$ and $\vcb_{t}^{(\ell)}$ are learned through training, while the values of $\fnrho_{t}$, $L_{t}$ and $N_{t}^{(\ell)}$ are design choices. In machine learning parlance, the former are known as parameters, while the latter are referred to as hyperparameters.

To learn the covariance matrix, we propose the following
\begin{equation} \label{eq:covLearn}
   \fnSigma_t(\vcx_{t-1},\vcy_{t}) = \, \fnSigma(\vcx_{t-1},\vcy_{t}) = \mtC \mtD(\vcx_{t-1},\vcy_{t}) \mtC^{\Tr}
\end{equation}
where $\mtC \in \fdR^{N \times N}$ and where $\mtD(\vcx_{t-1},\vcy_{t}) \in \fdR^{N \times N}$ is such that
\begin{equation} \label{eq:covKernel}
    \big[ \mtD(\vcx_{t-1},\vcy_{t}) \big]_{ij} = \exp \big( - ([\vcz_{t}]_{i} - [\vcz_{t}]_{j})^{2} \big)
\end{equation}
for $\vcz_{t} = \mathrm{NN}^{\Sigma}(\vcx_{t-1},\vcy_{t})$ is the output of a (different) neural network, yielding $\vcz_{t} \in \fdR^{N}$. Analogously to $\mathrm{NN}_{t}^{\mu}$, the learnable parameters of this neural network are given by the affine transformations at each layer, while the activation function, the number of layers, and the dimension of the hidden units are hyperparameters. The matrices $\mtC \in \fdR^{N \times N}$ are also learnable parameters, that are capable of learning different directions for the variance components. At this point, we note that each time instant represents an iteration of the SIS particle filter. Therefore, following \cite{Monga2021-AlgorithmUnrolling}, we propose to learn a separate neural network $\mathrm{NN}_{t}^{\mu}$ for the mean for each time instant [cf. \eqref{eq:NN}]. However, we opt to leave the learnable parameters of the covariance matrix be independent of time [cf. \eqref{eq:covLearn}].

A good sampling distribution is one where the resulting weights are all very similar to each other, leading to low degeneracy over time [cf. \eqref{eq:SISweights}]. Therefore, in order to train the neural networks and learn the corresponding parameters, we will maximize the following loss function
\begin{equation} \label{eq:lossFunction}
    \fnJ\big( \{\sctw_{t}^{(k)}\}_{t,k} \big) = \sum_{t=0}^{T-1} \sum_{k=1}^{K} \log \big( \sctw_{t}^{(k)} \big).
\end{equation}
One can check that~\eqref{eq:lossFunction} is maximized when all the weights are equal to $1/K$. 
Intuitively, maximizing \eqref{eq:lossFunction} implies that when weights $\sctw_{t}^{(k)}$ get too small, they get severely penalized by the logarithm and therefore the optimization algorithm will attempt to make them large again. 
On the other hand, the concavity of the logarithm reduces the benefit of further increasing a weight that is already large.
Hence, the overall effect is that the weights will tend to be close to $1/K$.

We emphasize that obtaining the parameters by maximizing \eqref{eq:lossFunction} results in the unsupervised learning of the sampling distribution. This has many advantages. First, it does not require access to the true trajectories $\{\vcx_{t}\}$ of the system (which are unobservable), only to the corresponding measurements $\{\vcy_{t}\}$. Second, it allows the sampling distribution to generalize to unseen trajectories (coming from the same system). Thus, unsupervised learning allows us to learn high-performing sampling distributions in a wider array of problems.

To finalize, we note that the weights $\sctw_{t}^{(k)}$ depend on the actual particles $\vcx_{0:t}^{(k)}$ that have been sampled from the distribution $\fnpi$ that we are learning. Therefore, in order to actually compute the gradient of $\fnJ$ with respect to our learnable parameters, we would require a way to propagate the gradient through the particles $\vcx_{0:t}^{(k)}$. We do so by means of the reparametrization trick, commonly used in training variational autoencoders~\cite{Kingma2014-VAE}. We observe in Sec.~\ref{sec:sims} that the proposed architecture successfully learns appropriate sampling distributions.




\section{Numerical Experiments} \label{sec:sims}



In this section, we consider the system model \eqref{eq:modelDescription} where only a sequence of measurements $\{\vcy_{t}\}$ is available. The goal is to obtain $\xp[\vcx_{t}|\vcy_{0:t}]$ as the estimator of the state $\vcx_{t}$ at time $t$ based on the measurements $\vcy_{0:t}$ from time $0$ to $t$. We present three different modeling scenarios to showcase the performance of the SIS particle filter with a sampling distribution learned as in Sec.~\ref{sec:unrolling}.

First, we consider a linear system with Gaussian noise (Sec.~\ref{subsec:linear}). In this scenario, the posterior distribution $\fnp(\vcx_{0:t}|\vcy_{0:t})$ is a multivariate normal so that the estimate $\xp[\vcx_{t}|\vcy_{0:t}]$ can be obtained in closed form. Also, the sampling distribution that minimizes weight degeneracy $\fnp(\vcx_{t}|\vcx_{t-1},\vcy_{t})$ is a multivariate normal that can be easily sampled. This scenario allows for an immediate comparison with two baselines that would be, otherwise, impossible to compute.

Second, we consider a nonlinear system with linear measurements and Gaussian noise (Sec.~\ref{subsec:nonlinear}). In this case, the posterior $\fnp(\vcx_{0:t}|\vcy_{0:t})$ cannot be obtained in closed form, but the sampling distribution that minimizes weight degeneracy $\fnp(\vcx_{t}|\vcx_{t-1},\vcy_{t})$ is still a multivariate normal \cite[eq. (13)]{Doucet2000-ParticleFiltering}. This scenario, while being more complicated than the previous one, still offers the minimum-degeneracy SIS filter as a benchmark.

Third, we consider a linear system with non-Gaussian noise (Sec.~\ref{subsec:nongaussian}). In this case, neither the posterior distribution nor the minimum-degeneracy SIS can be computed in closed form. However, we use this scenario to illustrate the robustness to model mismatch of the SIS particle filter with the learned sampling distribution. That is, we assume a Gaussian noise and we observe how the sampling distribution adapts to the actual, non-Gaussian model.

In all scenarios, we draw comparisons between the minimum-degeneracy SIS particle filter with a one where the distribution is learned as in Sec.~\ref{sec:unrolling}. We consider neural networks with 2 hidden layers, with output $256$ and $512$ features each. The chosen nonlinear activation function is the hyperbolic tangent. We train the models by running the particle filter, computing the loss function \eqref{eq:lossFunction}, and using ADAM \cite{Kingma15-ADAM} to update the parameters. The learning rate is $0.001$ and the forgetting factors are $0.9$ and $0.999$, respectively. We run the particle filter $200$ times, simulating $K=25$ particles each time, and updating the parameters after each of those runs.

For testing, we run the SIS particle filters $100$ times, drawing $K=25$ particles each time in order to estimate the target $\xp[\vcx_{t}|\vcy_{0:t}]$. We compute the relative root mean squared error (RMSE) between the estimate and the target value $\|\vchtheta_{t} - \vctheta_{t}\|_{2}/\|\vctheta_{t}\|_{2}$ where the estimate $\vchtheta_{t}$ is computed by means of \eqref{eq:particleFilter} for $\fnf_{t}(\vcx_{0:t}) = \vcx_{t}$, and where the target quantity is $\vctheta_{t}=\xp[\vcx_{t}|\vcy_{0:t}]$ for the first scenario, and $\vctheta_{t} = \vcx_{t}$ for the second and third scenarios (where $\xp[\vcx_{t}|\vcy_{0:t}]$ is not available in closed form). We run the entire training and testing $10$ times, randomly choosing the system model each time. We report the median and standard deviation of the relative RMSE.

We also consider particle filters with resampling \cite{Doucet2000-ParticleFiltering}. That is, when the value of $(\sum_{k=1}^{K}(\sctw_{t}^{(k)})^{2})^{-1} < K/3$, we sample the trajectories with replacement according to $\sctw_{t}^{(k)}$ resulting in more trajectories with higher $\sctw_{t}^{(k)}$, helping avoid degeneracy --at the expense of losing particle independence. This is a technique that is common practice when utilizing particle filters. We note that we do not use resampling during training, but only during testing.\footnote{The source code has been made available at \url{github.com/fgfgama/unrolling-particles}.}

\subsection{Linear system with Gaussian noise} \label{subsec:linear}

We start with the case of a linear model given by
\begin{equation} \label{eq:linearGaussianModel}
\begin{aligned}
    & \vcx_{t} = \mtA \vcx_{t-1} + \vcv_{t} \quad , \quad 
    \vcy_{t} = \mtC \vcx_{t}+ \vcw_{t}, \\
    & \vcx_{0} \sim \gaussDist(\vcmu^{0}, \mtSigma^{0}) \ , \ \vcv_{t} \sim \gaussDist(\vcZeros, \mtSigma_{v}) \ , \ \vcw_{t} \sim \gaussDist(\vcZeros, \mtSigma_{w}),
\end{aligned}
\end{equation}
where $\vcx_{t} \in \fdR^{N}$ and $\vcy_{t} \in \fdR^{M}$ so that $\mtA \in \fdR^{N \times N}$ and $\mtC \in \fdR^{M \times N}$. This implies that $\vcmu^{0} \in \fdR^{N}, \mtSigma^{0} \in \fdR^{N \times N}$, $\mtSigma_{v} \in \fdR^{N\times N}$, and $\mtSigma_{w} \in \fdR^{M \times M}$. We assume that the noise processes are white (i.e., $\vcv_{t}, \vcv_{t'}$ are uncorrelated for $t \neq t'$) and uncorrelated from each other. This model serves as a baseline, since all quantities of interest can be computed in closed form, namely the posterior distribution $\fnp(\vcx_{0:t}|\vcy_{0:t})$ and the optimal SIS particle filter $\fnp(\vcx_{t}|\vcx_{t-1},\vcy_{t})$.

We set $N = 10$, $M = 8$ and a trajectory that evolves for $12$ steps. We set $\mtA$ to be the adjacency matrix of a random planar graph with normalized spectral norm (i.e., it acts as a diffusion process) and we set $\mtC$ to be such that each measurement looks at the value of $\vcx_{t}$ at two different nodes and aggregates them, before adding the measurement noise. We set $\vcmu^{0} = \vcOnes_{10}$ and $\mtSigma^{0} = \mtI_{10}$, while the covariance matrices $\mtSigma_{v},\mtSigma_{w}$ are selected at random. We define the SNR as $\|\vcmu^{0}\|_{2}^{2}/\|\mtSigma_{v}\|_{2}^{2}$ and simulate for different values of it, ranging from $0\text{dB}$ to $10\text{dB}$. We set $\|\mtSigma_{v}\|_{2} = \|\mtSigma_{w}\|_{2}$.

\begin{figure*}[htb]
    \centering
    \subfloat[Linear]{%
    \label{subfig:linear}%
    \includegraphics[width=0.65\columnwidth]{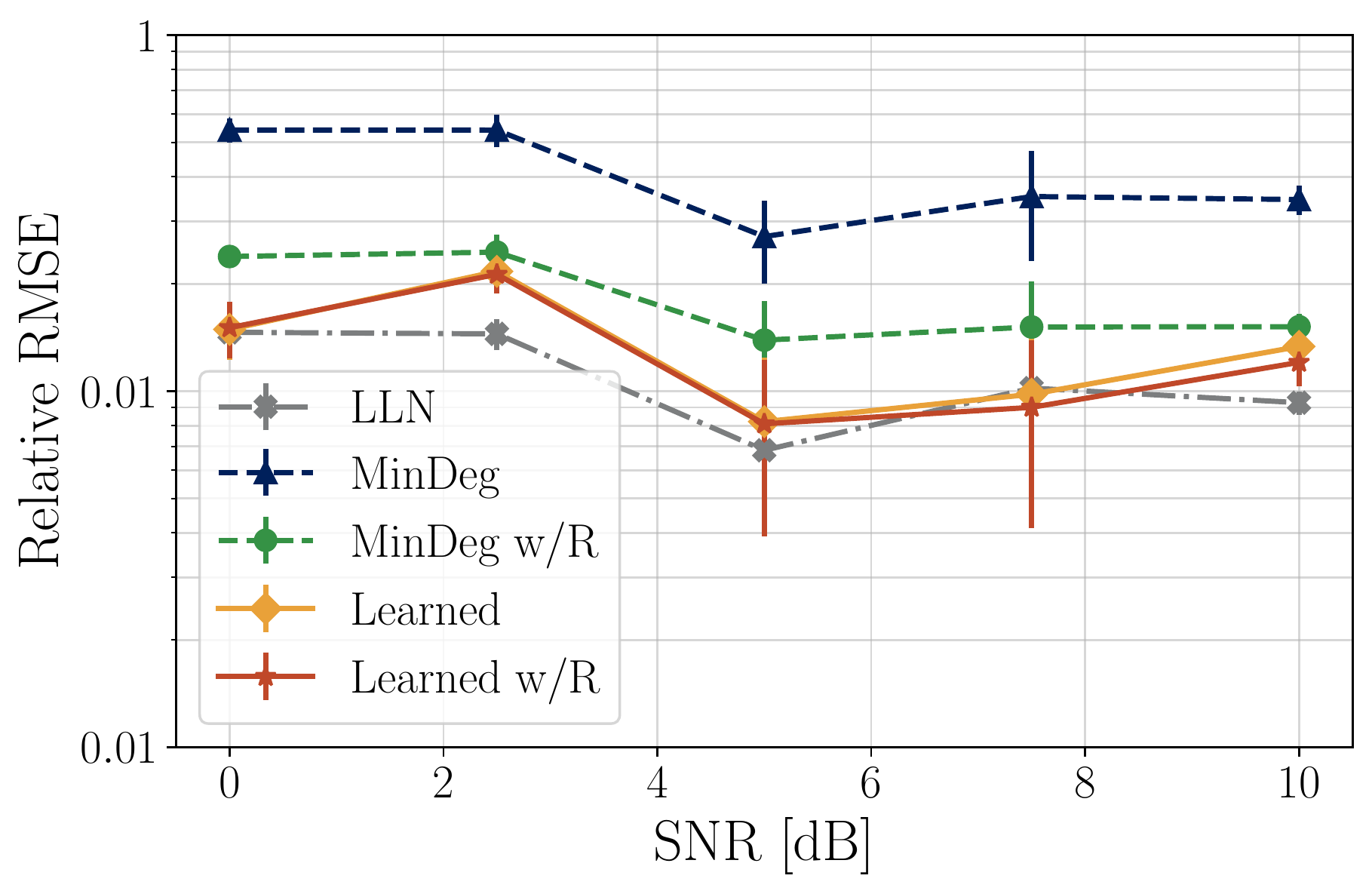}%
    }%
    \hfill
    \subfloat[Nonlinear]{%
    \label{subfig:nonlinear}%
    \includegraphics[width=0.65\columnwidth]{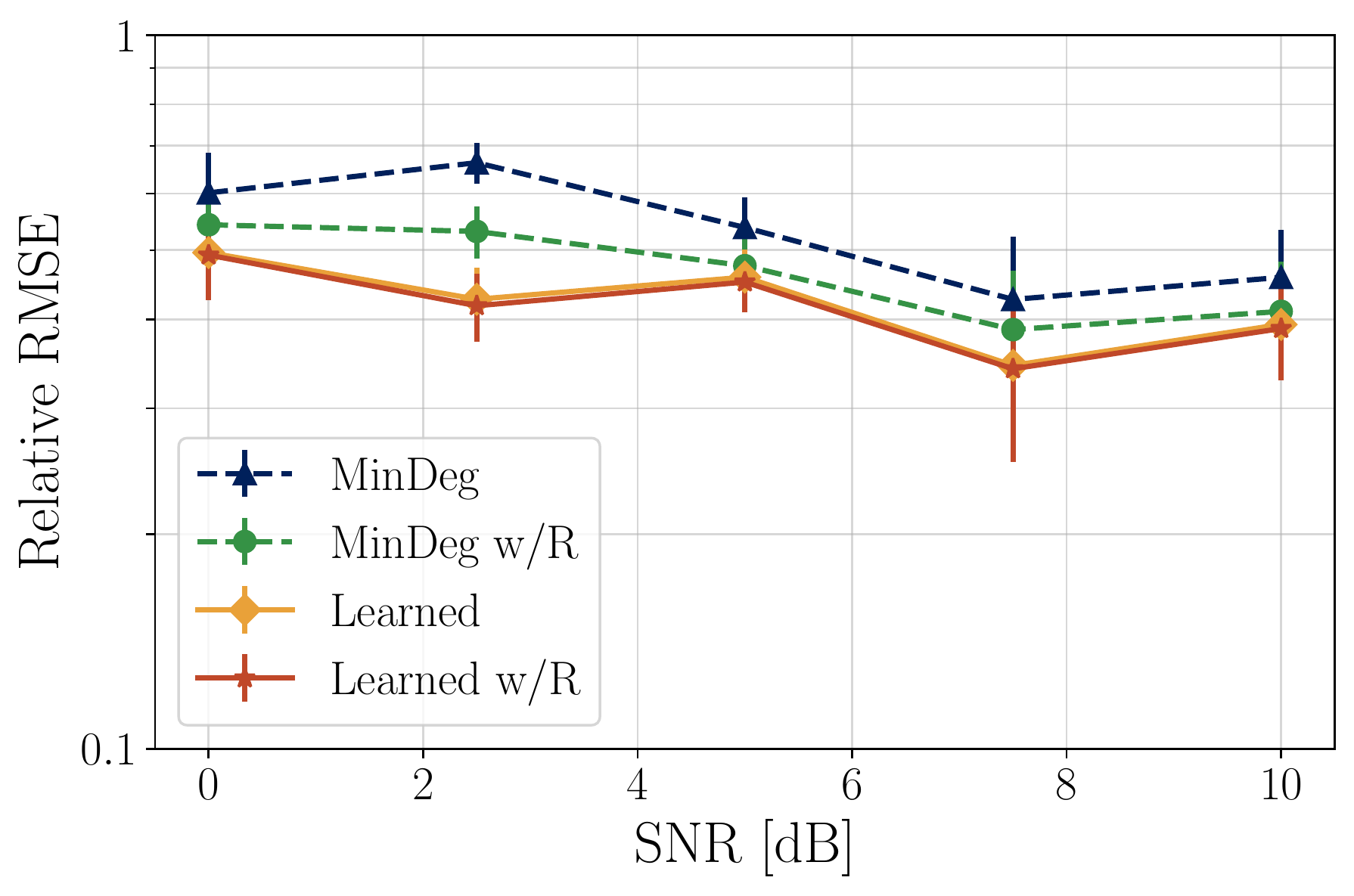}%
    }%
    \hfill
    \subfloat[Non-Gaussian]{%
    \label{subfig:nongaussian}%
    \includegraphics[width=0.65\columnwidth]{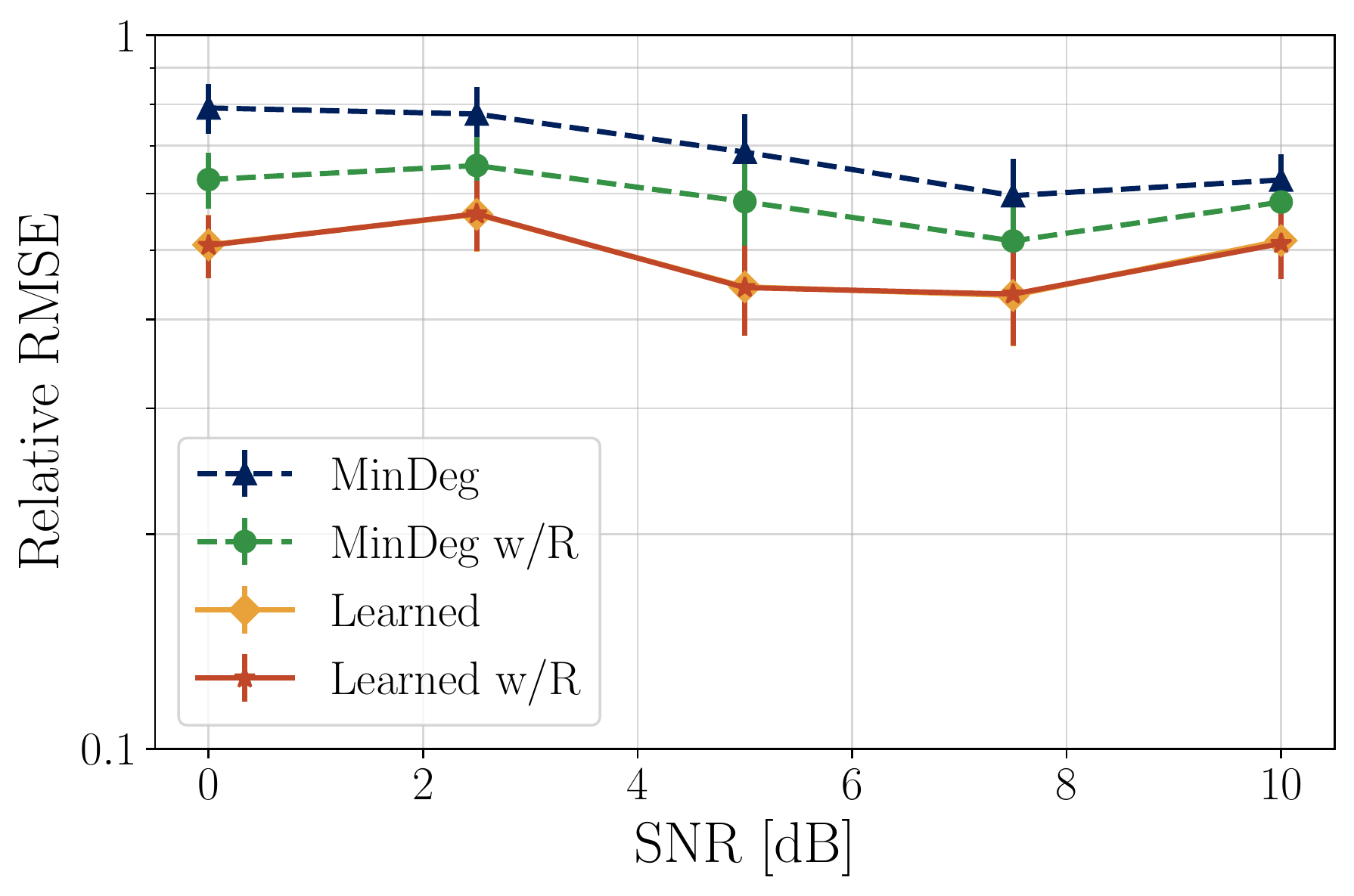}%
    }%
    \caption{Performance of the algorithms as a function of the SNR, measured by the relative RMSE. \protect\subref{subfig:linear}~Linear model with Gaussian noise. \protect\subref{subfig:nonlinear}~Nonlinear model with linear measurements and Gaussian noise. \protect\subref{subfig:nongaussian}~Linear model with uniform noise. In all cases it is observed that the learned SIS particle filter performs better. The error bars represent one-third of the estimated standard~deviation.}
    \label{fig:l2errorSNR}
\end{figure*}

We consider $5$ algorithms, namely: (i)~the application of the law of large numbers (LLN) sampling directly from $\fnp(\vcx_{0:t}|\vcy_{0:t})$ which serves as the baseline; (ii)~the minimum-degeneracy SIS particle filter sampling from $\fnp(\vcx_{t}|\vcx_{t-1},\vcy_{t})$ without resampling (iii)~and with resampling; (iv)~the SIS particle filter with learned distribution following Sec.~\ref{sec:unrolling}, without resampling, (v)~and with resampling. 

In Fig.~\ref{subfig:linear} we show the performance of the five algorithms as a function of the SNR. First, it is immediate that the learnable SIS particle filter exhibits better performance than the minimum-degeneracy (\texttt{MinDeg}) particle filter over the entire spectrum of SNR. Second, we note that while resampling significantly improves the \texttt{MinDeg} particle filter, it does not offer much improvement in the learnable case, likely because the distribution is trained to minimize degeneracy. Third, we note that the performance is overall robust, improving slightly for larger values of SNR. Fourth, we see that the performance is comparable to the LLN benchmark.

\vspace{0.4cm}

\subsection{Nonlinear system with Gaussian noise} \label{subsec:nonlinear}

For the second scenario, we add a nonlinear function $\fnphi:\fdR^{N} \to \fdR^{N}$ to the transition rule as
\begin{equation} \label{eq:nonlinearGaussianModel}
\begin{aligned}
    & \vcx_{t} = \fnphi(\mtA \vcx_{t-1}) + \vcv_{t}, \\ 
    & \vcy_{t} = \mtC \vcx_{t}+ \vcw_{t}, \\
    & \vcx_{0} \sim \gaussDist(\vcmu^{0}, \mtSigma^{0}) \ , \ \vcv_{t} \sim \gaussDist(\vcZeros, \mtSigma_{v}) \ , \ \vcw_{t} \sim \gaussDist(\vcZeros, \mtSigma_{w}),
\end{aligned}
\end{equation}
with all the other quantities defined analogously to Sec.~\ref{subsec:linear}. We choose the nonlinear function $\fnphi$ to be a pointwise absolute value, i.e. $[\fnphi(\vcx)]_{i} = |[\vcx]_{i}|$. We note that we cannot simulate the LLN estimate because we do not have access to the posterior $\fnp(\vcx_{0:t}|\vcy_{0:t})$. Recall that the minimum-degeneracy sampling distribution $\fnp(\vcx_{t}|\vcx_{t-1},\vcy_{t})$ is a multivariate normal with mean and variance given in \cite[eq. (13)]{Doucet2000-ParticleFiltering}, so that it can be easily sampled from it.

Results in Fig.~\ref{subfig:nonlinear} show that the learnable SIS particle filter performs better than the minimum-degeneracy one. We also see that while resampling helps the \texttt{MinDeg} filter, it does not make a difference for the learned filter. We also observe robustness with respect to varying the SNR, performing slightly better for larger values.

\vspace{0.4cm}

\subsection{Linear system with non-Gaussian noise: Model mismatch} \label{subsec:nongaussian}

For the last experiment we investigate the robustness to model mismatch. The model is given by \eqref{eq:linearGaussianModel}, except that now the initial state $\vcx_{0}$ as well as the noise $\vcv_{t}, \vcw_{t}$ are uniform, instead of Gaussian, with i.i.d. entries. The covariance matrices are given by $\mtSigma_{v} = \sigma^{2}\mtI_{10}$ and $\mtSigma_{w} = \sigma^{2}\mtI_{8}$ so that the SNR is $\|\vcmu_{0}\|_{2}^{2}/\sigma^{2}$.

To investigate model mismatch, we consider the minimum-degeneracy SIS filter assuming a multivariate normal distribution, given by the same sampling distribution $\fnp(\vcx_{t}|\vcx_{t-1},\vcy_{t})$ as in Sec.~\ref{subsec:linear}. Likewise, we do not modify the parametrization of the sampling distribution of Sec.~\ref{sec:unrolling} which is also given by a multivariate normal [cf. \eqref{eq:MVNormalParam}].

The results shown in Fig.~\ref{subfig:nongaussian} show the greater robustness of the learned distribution. In fact, in this scenario, the difference between the performance of the \texttt{MinDeg} SIS filter with resampling and the learned SIS filter is much larger than in the previous two scenarios.


\vspace{0.75cm}

\section{Conclusions} \label{sec:conclusions}



By leveraging neural networks together with algorithm unrolling, we proposed the first framework for learning particle filters in an unsupervised manner. Unsupervised learning allows us to leverage the measurements directly, avoiding the need for having access to the true trajectory and transferring better between training and testing. Future research directions of interest include: i)~looking at alternative parametrizations for the sampling distributions, especially focusing on multi-modal ones; ii)~analyzing joint learning of the distribution and the particle weights; and iii)~investigating distributed particle filtering approaches leveraging graph neural networks.

\vfill\pagebreak

\balance

\bibliographystyle{bibFiles/IEEEbib}
\bibliography{bibFiles/myIEEEabrv,bibFiles/biblioParticle}

\end{document}